\documentclass[11pt]{article}

\usepackage[final]{acl}

\usepackage{times}
\usepackage{latexsym}
\usepackage{makecell} % 

\usepackage[T1]{fontenc}
\usepackage[utf8]{inputenc}

\usepackage{microtype}

\usepackage{inconsolata}

\usepackage{graphicx}
\usepackage{amsmath}
\usepackage{booktabs}
\usepackage{multirow}

\usepackage{listings}
\title{AutoOntic: Automatic Ontology Induction from Text for Knowledge Graph Construction}

\author{
Alla Chepurova$^{1}$\quad\
Aydar Bulatov$^{1}$\quad
Mikhail Burtsev$^2$\quad
Yuri Kuratov$^{1}$\\\\
$^1$Cognitive AI Systems Lab, Moscow, Russia \\
$^2$London Institute for Mathematical Sciences, London, UK\\
\texttt{\{chepurova,bulatov,kuratov\}@cogailab.com}, \texttt{mb@lims.ac.uk}\\\\
}
\begin{document}
\maketitle
\begin{abstract}
Graph-based retrieval-augmented generation
depends on the quality of the knowledge graph
(KG) it retrieves over, yet schema-free LLM-
based extraction pipelines typically produce
flat graphs: deduplicated entities and relations
with no type hierarchy and no domain/range
constraints. We present AutoOntic, a pipeline
that automatically constructs an ontology from
text by canonicalizing open-vocabulary entity
types and relations, inducing a hierarchy over
the canonicalized types, and mining relation-
level domain and range constraints. Its hierarchy induction procedure prioritizes types using
corpus-derived relation-argument profiles and
places them through local, LLM-adjudicated
parent/child checks, allowing newly inserted
types to trigger reassignment of existing parent-
child links. The resulting hierarchy constrains
entity deduplication and supports the induction
of relation-level domain/range constraints. For
intrinsic evaluation, we introduce a reference-
based protocol that classifies induced ontology
types as exact, broader, narrower, or unrelated
to reference domain and range classes. On
the Wikidata-TekGen and DBpedia-WebNLG
subsets of Text2KGBench, over 90\% of Au-
toOntic’s induced endpoint types are semanti-
cally aligned with the reference classes across
two LLM backbones: about half match exactly,
while most remaining differences concern gran-
ularity. On MINE, knowledge graphs organized
using an ontology built by AutoOntic achieve
88\% retrieval accuracy. AutoOntic is available at \url{https://github.com/screemix/AutoOntic}.

\end{abstract}

\begin{figure}[t]
\centering
\includegraphics[height=0.46\textheight,keepaspectratio]{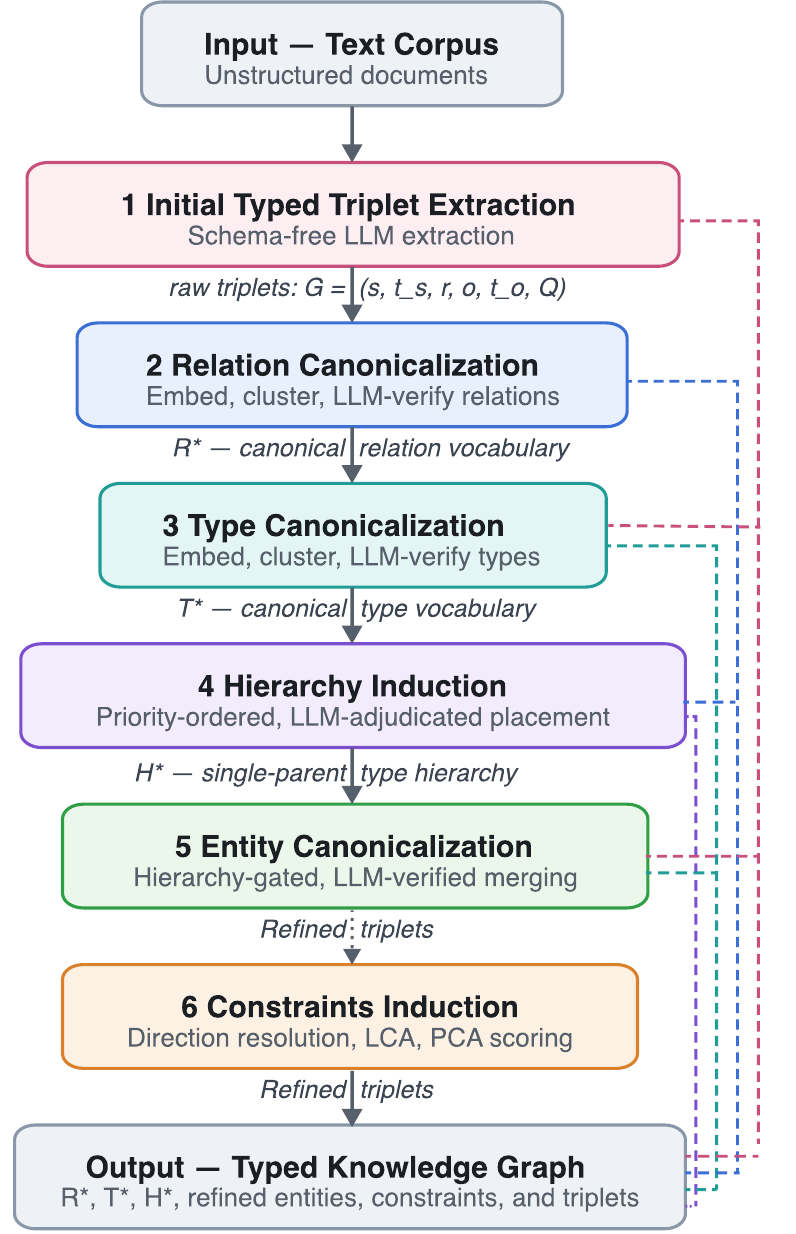}
\caption{The AutoOntic pipeline. A schema-free LLM extractor (\S\ref{sec:extraction}) produces a collection $\mathcal{G}$ of raw triplets $g=(s,t_s,r,o,t_o,Q)$. Their relation and type labels are canonicalized into a relation vocabulary $\mathcal{R}^*$ (\S\ref{sec:relation-dedup}) and a type vocabulary $\mathcal{T}^*$ (\S\ref{sec:type-dedup}) organized into a single-parent is-a hierarchy $\mathcal{H}^*$(\S\ref{sec:hierarchy-induction}). Entity mentions are then deduplicated using that hierarchy as a symbolic gate (\S\ref{sec:entity-dedup}), and domain/range constraints are mined by re-resolving $\mathcal{G}$ against $\mathcal{R}^*$, $\mathcal{T}^*$, and  $\mathcal{H}^*$(\S\ref{sec:constraints}). Colored dashed arrows mark additional direct inputs beyond the sequential chain.}
\label{fig:pipeline}
\end{figure}

\section{Introduction}
\label{sec:introduction}

Retrieval-augmented generation grounds large language model (LLM) outputs in retrieved evidence \citep{lewis2020retrieval}. However, retrieval over chunked text can struggle when answering multi-hop questions that require connecting evidence across documents. Graph-based approaches organize this evidence into knowledge graphs (KGs), where information is represented as entities and relations, enabling retrieval across connected facts \citep{edge2024graphrag}. The effectiveness of this approach depends not only on which facts a knowledge graph (KG) contains, but also on how those facts are organized. In particular, KG-RAG benefits from ontology \citep{chepurova2026wikontic}, a hierarchy of entity types that represent general categories and their specializations, and relation domain and range constraints that express which types of entities a relation connects. These structures provide a basis for category-level retrieval and structural validation, but must be constructed reliably to be useful.

Building a KG with this structure from text remains challenging. Schema-free extraction methods such as KGGen \citep{mo2025kggen} extract and canonicalize entities and relations to provide more compact and well-connected KG without requiring a predefined domain schema, but do not explicitly induce a hierarchy of entity types or relation domain and range constraints. Conversely, ontology-aware KG construction methods can use pre-built ontologies to organize and constrain extracted knowledge \citep{chepurova2026wikontic}, but applying an existing ontology to a new corpus may require aligning its vocabulary with corpus-specific expressions and introducing concepts absent from the original schema. These observations motivate a method that goes beyond flat KG construction to induce a corpus-grounded ontology, while retaining the option to incorporate an existing ontology when available.

We present \textbf{AutoOntic}, an ontology construction pipeline that extends schema-free approaches to close this gap: on top of a deduplicated entity and relation vocabulary, it produces a
hyponymy-aware, hierarchically organized type vocabulary, an entity
vocabulary deduplicated against that hierarchy, and a set of
per-relation domain/range constraints mined from it. The system can also extend an existing ontology when one is provided (\S\ref{sec:methods}). Our contributions are:
\begin{itemize}
  \item AutoOntic, a schema-free ontology induction method that combines corpus-derived relation-argument profiles with priority-ordered, top-down type placement. Unlike flat canonicalization, the procedure explicitly tests subsumption relationships and can retroactively reassign parent-child links when newly inserted types reveal a more specific placement. The resulting hierarchy is then used to constrain entity deduplication and to generalize observed relation domain/range types.
  \item An evaluation framework for schema-free ontology induction that assesses instance-level type alignment and induced domain/range constraints against a benchmark's reference ontology, distinguishing \textsc{Exact}, \textsc{More General}, and \textsc{More Narrow} matches (\S\ref{sec:evaluation}).

  \item Empirical evidence showing that AutoOntic's induced types largely align with reference ontologies, with disagreements predominantly reflecting differences in granularity rather than unrelated classes, and that its constructed knowledge graph achieves 85.87\% information-retention accuracy on MINE-1, rising to 88.00\% with Wikidata-seeded hierarchy initialization, outperforming KGGen (82.37\%) and Wikontic (82.93\%) (\S\ref{sec:results-text2kgbench}, \S\ref{sec:results-mine}).
\end{itemize}
\section{Methods: AutoOntic}
\label{sec:methods}

% \subsection{Overview}
% \label{sec:methods-overview}

AutoOntic turns a textual corpus into a canonicalized, typed knowledge graph: a deduplicated relation vocabulary, a deduplicated and hierarchically organized entity-type vocabulary, a deduplicated entity vocabulary with class assignments, and a set of relation-level domain/range constraints (Figure~\ref{fig:pipeline}).
At first, a schema-free LLM extractor  (\S\ref{sec:extraction}) produces a collection $\mathcal{G}$ of raw triplets,
each represented as:
\begin{equation}
g = (s, t_s, r, o, t_o, Q),
\end{equation}
where $s$ -- subject entity, $t_s$ -- subject type, $r$ -- relation predicate, $o$ -- object entity, $t_o$ -- object type, and $Q$ -- an optional list of qualifiers extracted from texts~\citep{chepurova2026wikontic}. Then $\mathcal{G}$ is passed further for canonicalization and typing. Every downstream stage resolves fields into canonical identifiers, keeping each stage's data dependencies explicit and independently inspectable.

% The pipeline runs ordered stages, described in
% \S\ref{sec:relation-dedup}--\S\ref{sec:constraints}:  \emph{(1) Initial typed triplet extraction}, \emph{(2) relation
% canonicalization}, \emph{(3) entity-type canonicalization}, which also
% induces a taxonomy hierarchy over the resulting type vocabulary,
% \emph{(4) entity class assignment and canonicalization}, and \emph{(5)
% domain/range constraint induction}. The ordering is dictated by data
% dependencies. Entity-type canonicalization is conditioned on the canonical relation vocabulary as a disambiguating signal. Entity deduplication runs after the type hierarchy because it uses the hierarchy as a symbolic gate for candidate synonyms. Constraint induction derives relation constraints directly from the extracted triplets and the outputs of the first two stages.

The AutoOntic pipeline comprises six ordered stages:
\emph{(1) typed triplet extraction, (2) relation canonicalization,
(3) entity-type canonicalization, (4) hierarchy induction,
(5) entity canonicalization and class assignment, and
(6) domain/range constraint induction}.
This ordering follows data dependencies: canonical relations inform
type-label disambiguation; the induced type hierarchy constrains
candidate entity merges; and the canonical vocabularies and
direction-corrected triplets provide evidence for domain/range constraints.

\paragraph{Shared canonicalization procedure.}
\label{sec:reoccurring-design}
The following design choices reoccur across stages. (1) Deduplication of synonymous surface forms is performed by clustering Contriever sentence embeddings \citep{izacard2022contriever} with HDBSCAN~\citep{campello2013hdbscan} at a stage-specific cosine-similarity threshold $\theta$, and every merge/split decision is adjudicated by a single LLM call per candidate cluster. (2) Merging is done through several rounds: each round re-embeds the current pool, re-clusters it, and sends multi-member clusters to the LLM for verification, folding accepted merges back into the pool for the next round. The procedure repeats until the vocabulary size stops reducing or a configurable round limit $R$ is reached. (3) The result of each stage is checkpointed once it completes, allowing independent inspection of every stage's output and reuse in follow-up experiments. Every $\theta$ and $R$ instance below is stage-specific (relation, type, and entity deduplication each set their own); concrete values for these and every other hyperparameter introduced in \S\ref{sec:relation-dedup}--\S\ref{sec:constraints} are collected in Table~\ref{tab:hyperparams} (Appendix~\ref{sec:appendix-hyperparams}).

\subsection{Typed Triplet Extraction}
\label{sec:extraction}
An LLM extracts subject--relation--object triplets enriched with subject/object types and contextual qualifiers with meta-information that cannot be expressed as a standalone triplet such as time, location, or condition. Types are initially unconstrained free text. Each extracted record has the form $g=(s,t_s,r,o,t_o,Q)$, where $s,o$ are entity mentions, $t_s,t_o$ their raw types, $r$ the relation, and $Q$ an optional qualifier list. The extracted records form $\mathcal{G}$. The full extraction prompt is given in Appendix~\ref{sec:appendix-prompts}, Listing~\ref{lst:extraction}.

\subsection{Relation Canonicalization}
\label{sec:relation-dedup}

Relation canonicalization maps the raw, free-text \texttt{relation} strings observed across all triplets onto a canonical relation vocabulary
$\mathcal{R}^*$. Candidate clusters are generated by embedding every raw
relation surface form and clustering with HDBSCAN at similarity threshold
$\theta_r$, as described in \S\ref{sec:reoccurring-design}. Every multi-member candidate cluster is then passed to a single LLM call that decides whether to merge all members into one relation or split them into several (see the prompt in Listing~\ref{lst:cluster-relations}); this repeats for up to $R_r$ rounds until no round shrinks $\mathcal{R}^*$ any further.

\subsection{Entity-Type Canonicalization}
\label{sec:type-dedup}

Raw type labels are canonicalized into $\mathcal{T}^*$.
Canonical types are then organized into a single-parent hierarchy
$\mathcal{H}^*$, which may form a forest.
Here, entity types correspond to ontology classes.

\paragraph{Type Deduplication}
Candidate clusters over raw type labels are generated the same way as for
relations (\S\ref{sec:relation-dedup}, threshold $\theta_t$), but with one additional source of evidence: each candidate label shown to the LLM verifier is annotated with the top $k_{\text{ctx}}$ canonical relations it most often co-occurs with (a PPMI-weighted relation-argument profile). These profiles provide corpus-derived structural evidence beyond lexical or embedding similarity: two types may have similar labels but different relational roles, or different labels but nearly identical distributions of relational contexts. The verification step uses this evidence alongside the labels to decide whether a candidate pair should be merged.

The verification prompt additionally enforces a hyponymy (``kind of'') test: two labels are merged only if neither is a hyponym of the other, even when they are lexically similar or their relation-argument profiles overlap almost entirely. For example, \ \emph{documentary film} and \emph{film} are similar and participate in the same relations, but LLM is instructed to keep them distinct, whereas \emph{movie} and \emph{film}, true synonyms, should
be merged. The full prompt is given in Listing~\ref{lst:cluster-entity-types}. As in relation deduplication, this embed-cluster-verify cycle repeats for up to $R_t$ rounds until a round adds no further reduction to $\mathcal{T}^*$.

\subsection{Hierarchy Induction}
\label{sec:hierarchy-induction}
Flat type canonicalization removes surface redundancy but does not encode subsumption relationships between them. AutoOntic therefore induces an explicit is-a hierarchy over $\mathcal{T}^*$ to support two operations that depend on these relationships: restricting entity-merge candidates to structurally compatible types
(\S\ref{sec:entity-dedup}) and generalizing observed domain/range types to shared ancestors when estimating relation constraints (\S\ref{sec:constraints}).

In AutoOntic, canonical types are organized into a single-parent hierarchy $\mathcal{H}^*$, which may form a forest. Rather than asking an LLM to organize the entire type vocabulary in a single call, AutoOntic decomposes placement into local decisions over the current hierarchy $\mathcal{H}^*$, inserting types one by one from general to specific. To find such ordering of entity types, we compare the PPMI-weighted relation-argument profiles associated with each deduplicated type. Following the distributional inclusion principle, a more general type is expected to occur in a broader range of relational contexts than its more specific subtypes. We use Weeds precision \citep{weeds2004characterising} to estimate the extent to which the contexts associated with one type are contained in those associated with another. The resulting containment scores define an approximate generality score based on how many other types could fit beneath each type. Types are processed in descending score order; these scores are not treated as proof of subsumption. Actual parent-child relationships are determined by the subsequent LLM-adjudicated placement procedure, which can revise earlier attachments as additional types are inserted. The hierarchy can be initialized from a seed or existing ontology and extended with types discovered in new text. Selected seed categories can be pinned as roots.

For each type, AutoOntic traverses the hierarchy downward from the current roots. At each level, the LLM performs separate \textbf{parent} and, if no parent is found, \textbf{child} checks. The parent check asks whether an existing candidate subsumes the new type. A parent match continues traversal through that node's children; a synonymous candidate is instead merged with the new type. The child check asks whether the new type subsumes any existing candidates. A child match attaches the matched nodes beneath the new type and stops traversal. If neither succeeds, the type is inserted at the current position. When a level contains more than $B$ candidates, a consolidation call attaches clear siblings under synthesized parents proposed by the LLM.

This design keeps each decision focused on a bounded part of the hierarchy, reducing the context required for an individual LLM call and avoiding repeated presentation of the entire ontology. More importantly, the decisions have explicit structural consequences: a later, more general type can correct an earlier attachment by becoming the parent of an existing node. Thus, the procedure does not merely append labels to a taxonomy; it incrementally constructs and repairs the hierarchy as new evidence becomes available. Placement depends on the existing hierarchy and the order in which types are processed. The local checks, consolidation procedure, and reassignment rules are specified in Appendix~\ref{sec:appendix-prompts}. Figure~\ref{fig:hierarchy-steps} demonstrates the hierarchy induction algorithm. Figure~\ref{fig:hierarchy-steps-example} shows the more detailed version of the algorithm step-by-step on a progressively populated movie ontology.

\begin{figure}[t]
\includegraphics[width=\columnwidth]{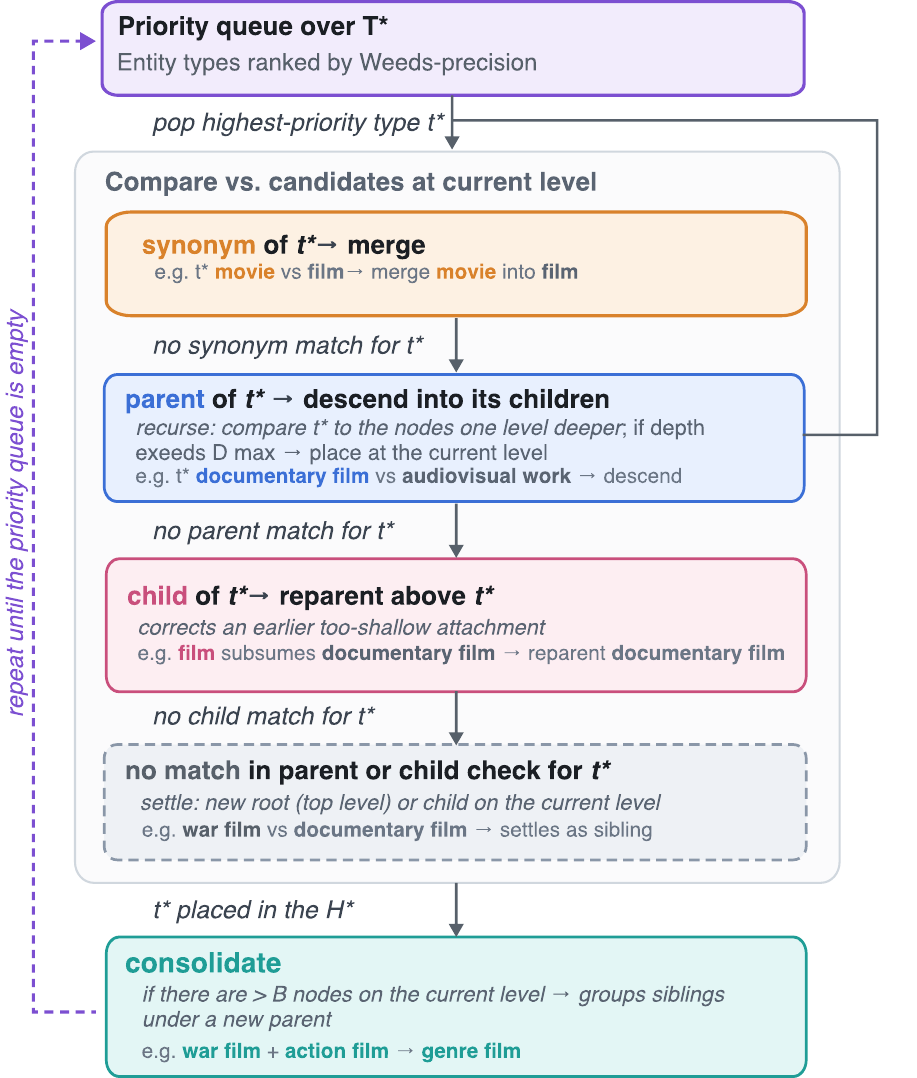}
\caption{Hierarchy induction algorithm. Canonical types in $\mathcal{T}^*$ are processed in descending order of a generality score based on Weeds precision over their relation-argument profiles. Starting from the roots of $\mathcal{H}^*$, an LLM first checks whether an existing candidate is a parent of the current type $t\in\mathcal{T}^*$. Synonymous types are merged; otherwise, a parent match continues traversal through that candidate's children. If no parent is found, a separate child check identifies existing candidates to attach beneath $t$. If no match is found, $t$ is added as a sibling at the current level. When the number of nodes at a level exceeds threshold $B$, an LLM consolidates related siblings under a synthesized parent, limiting the context required for subsequent placement decisions. The procedure repeats until all types have been processed.}
\label{fig:hierarchy-steps}
\end{figure}

% \begin{figure*}[t]
% \centering
% \includegraphics[width=\textwidth]{hierarchy_induction.pdf}
% \caption{Hierarchy induction on a film-domain vocabulary, one placement decision per step. (a) The initial priority ranking; (b)-(c) The first two nodes settle with \emph{documentary film} attached directly under \emph{audiovisual work} -- too shallow, since \emph{film} does not exist yet. (d) Once \emph{film} is placed, it judges itself \textbf{parent of} \emph{documentary film}: the earlier edge is removed and \emph{documentary film} is reattached beneath \emph{film}; (e) \emph{movie} merges into \emph{film} as a \textbf{synonym}. (f) \emph{war film} settles under \emph{film} \textbf{as a child}; (g) \emph{action film} is judged \textbf{same class} as \emph{war film}: a new shared parent, \emph{genre film}, is synthesized, and \emph{war film}'s edge moves from \emph{film} to it; (h) \emph{biographical film} settles under \emph{documentary film}, giving the final tree.}
% \label{fig:hierarchy-steps}
% \end{figure*}

\paragraph{Depth bound and guarantees.} A maximum-depth cap $D_{\max}$ bounds the worst-case number of placement decisions for any candidate node. A node that would need to exceed it is placed into a structurally separate unresolved bucket rather than being force-attached or discarded, allowing to extent the resulted hierarchy later on. The result of such procedure is always a single-parent forest at every
point in the procedure: acyclicity and the absence of transitive shortcut
edges follow directly from the construction.

\subsection{Entity Canonicalization and Class Assignment}
\label{sec:entity-dedup}

% Entity canonicalization step assigns each entity its canonical type  and deduplicates synonymous mentions. An entity's identity is the compound pair (name, canonical type) rather than name alone, so that homonymous names typed differently (e.g.\ \emph{Paris} the city versus \emph{Paris} the person) are never conflated by construction.

Entity mentions are represented by $(\text{name},\text{type})$, allowing identical names with different types to remain distinct (e.g., \emph{Paris} the city and \emph{Paris} the person).

% Before embedding-based comparison, candidate pairs are restricted by a symbolic gate: two mentions are only ever compared if they share literally the same type, or are siblings under the same parent in the induced type hierarchy (\S\ref{sec:hierarchy-induction}).
Initial comparisons are restricted to mentions with the same type or sibling types under the same immediate parent in $\mathcal{H}^*$ (\S\ref{sec:hierarchy-induction}).
Clustering entities surface forms is performed with HDBSCAN  among candidates within one sibling
group. Within each such partition, merging then follows the round-based procedure common to every deduplication stage (\S\ref{sec:reoccurring-design}, Listing~\ref{lst:cluster-entity-names}), applied independently per partition. Candidates are clustered at similarity threshold $\theta_e$, and each partition's embed-cluster-verify cycle repeats for up to $R_e$ rounds.

A final reconciliation pass broadens candidate comparisons across sibling groups whose types share a common ancestor in $\mathcal{H}^*$. A merged entity retains the union of its mention types, with their lowest common ancestor assigned as its primary type.

\subsection{Domain/Range Constraint Induction}
\label{sec:constraints}

Constraint induction first re-resolves every raw triplet against the canonical relation and type vocabularies, $\mathcal{R}^*$ and $\mathcal{T}^*$, produced by \S\ref{sec:relation-dedup} and \S\ref{sec:type-dedup}. Because relation canonicalization may merge a predicate with its grammatical inverse (e.g.\ \emph{directed} and \emph{was directed by}), each canonical relation with multiple surface forms is processed by an LLM that classifies each form as \emph{forward} or \emph{inverted}; inverted forms then have their subject and object types swapped before constraint estimation.

For each canonical relation, domain and range types are considered independently. We compute the exact lowest common ancestor (LCA) of observed types in $\mathcal{H}^*$. If observations span disconnected roots, they are partitioned by root and an LCA is computed within each partition, yielding multiple domain/range signatures when a relation connects unrelated type families.

Finally, triplets are grouped by their direction-corrected $(T_d,T_r)$ signature, and each signature is scored by a confidence equal to the share of relation $r$'s own triplets it accounts for:
\begin{equation}
\mathrm{conf}(r,T_d,T_r)=
\frac{\mathrm{support}(r,T_d,T_r)}{\mathrm{total}(r)},
\end{equation}
where $T_d$ and $T_r$ are the domain and range entity types, respectively.
% Constraints are assigned \textsc{Hard}, \textsc{Soft}, or \textsc{Hint} strength by thresholding it at $\tau_{\mathrm{hard}}$, $\tau_{\mathrm{soft}}$, and $\tau_{\mathrm{hint}}$.
% \section{KG Retrieval and QA}
% TODO -- not yet written.

\section{Evaluation}
\label{sec:evaluation}

\subsection{Ontology evaluation}
\label{sec:onto-eval}
\textbf{Benchmark}. We evaluate AutoOntic on \textbf{Text2KGBench}
\citep{mihindukulasooriya2023text2kgbench}, a benchmark of sentence--triple
pairs annotated against small, hand-curated domain ontologies, comprising
two subsets: \textbf{Wikidata-TekGen} (10 domain ontologies) and
\textbf{DBpedia-WebNLG} (19 domain ontologies). Each domain ontology
defines a set of relations coupled with domain and range types (allowed subject and object types respectively). Every test sentence is
paired with a gold triple set $\{(s, r, o)\}$, where $r$ is drawn from the ontology's
relation vocabulary. We use the \texttt{test} split with 4{,}062 sentences across the 10 Wikidata-TekGen ontologies and 2{,}014 across the 19 DBpedia-WebNLG ontologies. 
% Gold triplets carry no per-triple type annotation, so a relation's declared domain/range class stands in for the type of subject/object. 

% Several Wikidata-TekGen ontologies
% additionally encode a class hierarchy via \texttt{rdfs:subClassOf}, whereas
% the DBpedia-WebNLG ontologies did not.

Text2KGBench's intended task is \emph{ontology-guided} generation: a system is given the ontology and must extract compliant triplets. We consider the inverse setting: AutoOntic is schema-free and receives no ontology as input. Instead, it must induce a schema and extract triplets from the text. We then compare the induced schema and extracted triplets against the benchmark's gold ontology and gold triplets.

We used two LLM backbones driving every stage of the pipeline: Qwen3-Next-80B-A3B-Instruct
\citep{qwen2025qwen3} and gpt-oss-120b \citep{openai2025gptoss}, both run with the identical pipeline hyperparameters listed in Table~\ref{tab:hyperparams} (Appendix~\ref{sec:appendix-hyperparams}). Judging is performed in every condition by a fixed model, GPT-5.1, independent of whichever backbone built the KG. 

% We run
% two conditions per benchmark subset:

% \begin{description}
%   \item[\textsc{Separate}] one pipeline run per ontology domain, yielding
%   an independent $\mathcal{R}^*$, $\mathcal{T}^*$, and hierarchy for that
%   domain alone.
%   \item[\textsc{Combined}] one pipeline run over the Averaged sentences of
%   every domain in the subset, yielding a single $\mathcal{R}^*$,
%   $\mathcal{T}^*$, and hierarchy shared across all of them.
% \end{description}

% The \textsc{Combined} condition tests whether schema-free discovery
% separates unrelated domains into distinct regions of one induced hierarchy
% on its own, rather than only ever being evaluated one domain at a time.

\textbf{Triple matching.} For each gold triple, we collect the triples extracted by AutoOntic from the same sentence and ask an LLM judge to identify an extracted triple that expresses the same underlying fact, allowing for paraphrased relation labels and inverse formulations that swap the subject and object (e.g., \emph{X composed Y} vs.\ \emph{Y was composed by X}). The judge returns the index of a matching triple and labels its direction as \emph{same} or \emph{inverse}, or indicates that no match exists. Matching is one-to-one: once an extracted triple is matched to a gold triple, it is removed from the candidate pool and cannot be matched to another gold triple.

We match triples from the gold side outward to evaluate recovery of the reference annotations without automatically treating every unmatched extraction as an error. Text2KGBench uses curated relation vocabularies containing roughly 15--20 relations per domain, whereas a schema-free extraction system may recover valid facts that are not represented in the gold annotations. We therefore report three complementary metrics. \textbf{Recall} is the proportion of gold triples matched by extracted triples. The \textbf{extra rate} is the proportion of extracted triples left unmatched after all gold triples have been processed; these triples may represent facts omitted from the reference annotations rather than extraction errors. We report the extra rate separately from recall because unmatched extractions cannot be reliably classified as incorrect solely on the basis of their absence from the gold annotations.
% Finally, the \textbf{direction breakdown} reports the proportions of matched triples classified as \emph{same} and \emph{inverse}, providing a diagnostic of relation-direction consistency. 

\textbf{Domain/Range Type Matching.} For each matched triple, we compare AutoOntic's subject and object types with the gold relation's domain and range classes. For matches classified as \emph{inverse}, we swap the gold domain and range. An LLM judge classifies each comparison as \textsc{Exact} (equivalent types), \textsc{More General} (AutoOntic's type subsumes the gold type), \textsc{More Narrow} (AutoOntic's type is a subtype of the gold type), or \textsc{Not Related}. We report the distribution of these judgments separately for domain and range, thereby distinguishing exact type alignment in triplets from differences in the generality of the induced types.

\textbf{Constraint Plausibility.} Type matching evaluates individual triples, whereas AutoOntic also induces relation-level domain and range constraints by generalizing observed argument types using their lowest common ancestor (LCA). To evaluate these constraints, for each matched triple we identify the induced constraint(s) whose domain and range cover the triple's subject and object types, accounting for inverse matches and allowing multiple constraints when a relation spans disconnected hierarchy partitions. We then compare each selected constraint's domain and range with the corresponding gold classes using the same four judgment categories as in Domain/Range Type Matching. We report the resulting distributions alongside each constraint's support and confidence, including constraints below the $\tau_{\text{hint}}$ threshold and regardless of their \textsc{Hard}, \textsc{Soft}, or \textsc{Hint} tier (\S\ref{sec:constraints}). Comparing instance-level type judgments with constraint-level judgments shows how well the induced constraints align with the reference ontology and whether they tend to be more general or more specific than the gold classes.

Text2KGBench's Ontology Conformance metric measures whether extracted triples use relations from the input ontology's canonical relation vocabulary; leaving domain/range validation as a direction for future extension~\citep{mihindukulasooriya2023text2kgbench}. Our evaluation complements this metric by assessing domain and range alignment at both the instance and induced-constraint levels, using graded judgments that distinguish equivalent, broader, narrower, and unrelated types.

\subsection{Information Retention}
\label{sec:eval-mine}

\begin{table*}
\centering
\small
\resizebox{\textwidth}{!}{%
\begin{tabular}{llll cccc cccc}
\toprule
& & & & \multicolumn{4}{c}{Type Match} & \multicolumn{4}{c}{Constraint Plausibility} \\
\midrule
Dataset & Condition & Model & Side & Exact & \makecell{More\\General} & \makecell{More\\Narrow} & Sum & Exact & More Gen. & \makecell{More\\Narrow} & Sum \\
\midrule
\multirow{16}{*}{\makecell{Wikidata \\ TekGen}}
 & \multirow{4}{*}{--} & \multirow{2}{*}{Qwen3-Next} & D & 58.6 & 14.7 & 17.5 & 90.8 & 9.6  & 34.9 & 1.6 & 46.1 \\
 & & & R  & 53.6 & 22.6 & 16.3 & 92.5 & 12.2 & 33.3 & 2.6 & 48.1 \\
 & & \multirow{2}{*}{gpt-oss} & D & 58.0 & 10.9 & 23.1 & 92.0 & 9.1  & 53.9 & 1.2 & 64.2 \\
 & & & R  & 57.2 & 22.1 & 14.6 & 93.9 & 12.3 & 30.7 & 3.7 & 46.7 \\
\cmidrule(lr){2-12}
 & \multirow{4}{*}{\makecell{Wikidata \\ Seeded}} & \multirow{2}{*}{Qwen3-Next} & D & 58.5 & 14.6 & 17.6 & 90.7 & 3.4  & 77.7 & 1.6 & 82.7 \\
 & & & R  & 53.8 & 22.8 & 16.3 & 92.9 & 5.1  & 49.9 & 1.2 & 56.2 \\
 & & \multirow{2}{*}{gpt-oss} & D & 57.6 & 11.1 & 23.0 & 91.7 & 4.6  & 41.2 & 3.1 & 48.9 \\
 & & & R  & 57.9 & 21.7 & 13.5 & 93.1 & 5.3  & 35.4 & 3.2 & 43.9 \\
\cmidrule(lr){2-12}
 & \multirow{4}{*}{$D \leq 5$} & \multirow{2}{*}{Qwen3-Next} & D & 58.2 & 14.9 & 18.0 & 91.1 & 3.9  & 76.3 & 3.3 & 83.5 \\
 & & & R  & 54.5 & 22.0 & 16.3 & 92.8 & 13.0 & 63.6 & 1.2 & 77.8 \\
 & & \multirow{2}{*}{gpt-oss} & D & 57.5 & 10.9 & 22.8 & 91.2 & 9.7  & 58.0 & 2.5 & 70.2 \\
 & & & R  & 57.5 & 21.0 & 14.0 & 92.5 & 14.4 & 43.6 & 9.7 & 67.7 \\
\cmidrule(lr){2-12}
 & \multirow{4}{*}{$D \leq 10$} & \multirow{2}{*}{Qwen3-Next} & D & 58.7 & 14.4 & 17.9 & 91.0 & 4.5  & 91.0 & 1.7 & 97.2 \\
 & & & R  & 53.8 & 22.5 & 16.7 & 93.0 & 11.7 & 82.0 & 1.9 & 95.6 \\
 & & \multirow{2}{*}{gpt-oss} & D & 57.6 & 10.8 & 23.2 & 91.6 & 7.4  & 88.2 & 1.4 & 97.0 \\
 & & & R  & 57.2 & 21.7 & 13.5 & 92.4 & 12.2 & 63.4 & 2.4 & 78.0 \\
\midrule
\multirow{16}{*}{\makecell{DBpedia \\ WebNLG}}
 & \multirow{4}{*}{--} & \multirow{2}{*}{Qwen3-Next} & D & 33.5 & 37.3 & 17.6 & 88.4 & 18.5 & 60.1 & 5.0 & 83.6 \\
 & & & R  & 39.8 & 25.0 & 23.9 & 88.7 & 21.1 & 49.1 & 13.8 & 84.0 \\
 & & \multirow{2}{*}{gpt-oss} & D & 27.3 & 36.7 & 23.8 & 87.8 & 21.3 & 56.9 & 6.1 & 84.3 \\
 & & & R  & 42.4 & 17.0 & 28.6 & 88.0 & 35.8 & 42.7 & 8.7 & 87.2 \\
\cmidrule(lr){2-12}
 & \multirow{4}{*}{\makecell{Wikidata \\ Seeded}} & \multirow{2}{*}{Qwen3-Next} & D & 33.0 & 37.1 & 17.4 & 87.5 & 9.5  & 78.1 & 3.8 & 91.4 \\
 & & & R  & 39.7 & 23.7 & 24.9 & 88.3 & 6.7  & 63.9 & 4.7 & 75.3 \\
 & & \multirow{2}{*}{gpt-oss} & D & 27.2 & 36.6 & 23.5 & 87.3 & 14.4 & 54.2 & 6.2 & 74.8 \\
 & & & R  & 42.1 & 16.7 & 28.8 & 87.6 & 22.2 & 47.7 & 8.3 & 78.2 \\
\cmidrule(lr){2-12}
 & \multirow{4}{*}{$D \leq 5$} & \multirow{2}{*}{Qwen3-Next} & D & 33.2 & 37.3 & 17.5 & 88.0 & 11.8 & 53.1 & 5.5 & 70.4 \\
 & & & R  & 39.4 & 23.6 & 25.0 & 88.0 & 11.6 & 43.6 & 9.3 & 64.5 \\
 & & \multirow{2}{*}{gpt-oss} & D & 27.2 & 36.5 & 23.6 & 87.3 & 17.3 & 60.4 & 4.1 & 81.8 \\
 & & & R  & 42.2 & 16.5 & 28.9 & 87.6 & 28.6 & 51.1 & 7.9 & 87.6 \\
\cmidrule(lr){2-12}
 & \multirow{4}{*}{$D \leq 10$} & \multirow{2}{*}{Qwen3-Next} & D & 33.2 & 37.3 & 17.5 & 88.0 & 18.3 & 38.7 & 6.4 & 63.4 \\
 & & & R  & 39.2 & 24.2 & 25.1 & 88.5 & 13.4 & 34.5 & 16.9 & 64.8 \\
 & & \multirow{2}{*}{gpt-oss} & D & 27.1 & 36.6 & 23.5 & 87.2 & 20.6 & 57.3 & 5.5 & 83.4 \\
 & & & R  & 41.9 & 17.0 & 29.1 & 88.0 & 32.1 & 46.2 & 6.6 & 84.9 \\
\bottomrule
\end{tabular}%
}
\caption{Per-triplet type match and Constraint plausibility of discovered ontology, reported separately for relation domains (\textit{D}) and ranges (\textit{R}) (\%).
\textit{Type Match} compares each triplet's domain/range types with the gold type and shows the share that matches it exactly, is more general (an ancestor), or is more narrow (a descendant).
\textit{Constraint Plausibility} applies the same breakdown to the induced relation constraints.
\textit{Sum} is the total of the three categories; the remainder up to 100 corresponds to types or constraints with no hierarchical relation to the gold ones.
Conditions: -- = unconstrained discovery, Wikidata-Seeded = discovery initialised with seed classes from the Wikidata ontology, $D \leq k$ = class hierarchy depth limited to $k$.
Type Match is nearly invariant across conditions, whereas Constraint Plausibility shows variation depending on conditions and quantitatively characterizing the underlying ontology.}
\label{tab:text2kgbench-types}
\end{table*}

\textbf{Dataset}. We complement reference-ontology alignment \S\ref{sec:onto-eval} with MINE-1 \citep{mo2025kggen}, which evaluates how much factual information from source text remains recoverable from a constructed KG. This directly tests a practical requirement of ontology-aware KG construction: organizing and canonicalizing extracted knowledge should preserve the information needed for subsequent retrieval.
We use all 100 essays (topic and content) and their
associated generated factual queries (1{,}500 in total). Every essay's triplets are extracted
(\S\ref{sec:extraction}) into one combined corpus and passed through the
pipeline once, yielding a single shared ontology and KG spanning all 100 essays. We evaluate both construction without a supplied ontology and a Wikidata-seeded variant, whose type hierarchy is initialized with Wikidata classes. This comparison assesses whether existing ontology structure can further improve information retention.

\textbf{Task and method.} We mirror KGGen's own MINE evaluation methodology exactly, for direct comparability to its reported numbers on the same benchmark: each generated query is both the retrieval query and the fact to check for. Context is retrieved from AutoOntic's constructed KG via KGGen's own
method (sentence-embedding similarity over entities
followed by a fixed-depth graph walk, top-$k=8$), and an LLM judge is asked
a single binary question: whether the retrieved context contains the
information stated in the query. We report \emph{overall accuracy}: the fraction of the
1{,}500 queries the judge scores as supported by their retrieved context,
both averaged over the whole benchmark and broken down per essay.

% \textbf{Results.} Table~\ref{tab:mine} reports overall accuracy: gpt-oss-120b reaches 73.93\% (1{,}109/1{,}500),
% judged independently by GPT-4o; gpt-4o itself reaches 62.53\%
% (938/1{,}500). Unlike the Text2KGBench evaluation above, the gpt-4o row is
% \emph{not} judged independently of its own construction backbone -- GPT-4o
% is both the KG-building model and the judge for that row, since MINE's
% judge is fixed to GPT-4o throughout. We found gpt-4o's lower score traces
% to extraction, not to canonicalization or retrieval: on this corpus
% gpt-4o's single-call extraction consistently yields markedly fewer raw
% triplets per document than gpt-oss-120b's under an identical prompt (a
% gap we confirmed holds across multiple gpt-4o access paths and is not an
% artifact of chunking, parsing, or truncation), leaving a sparser KG for
% the same retrieval procedure to draw on.

\begin{table}[t]
\centering
\small
\resizebox{\columnwidth}{!}{
\begin{tabular}{lc}
\toprule
Method & MINE-1 score (\%) \\
\midrule
AutoOntic, gpt4.1, Wikidata-Seeded & \textbf{88.00} \\
AutoOntic, gpt4.1 & 85.87 \\
AutoOntic, gpt4.1, $D\leq5$ & 85.27 \\
AutoOntic, gpt4.1, $D\leq10$ & 85.47\\
KGGen, gpt4.1 & 82.37\\
Wikontic, gpt4.1 & 82.93 \\
% GraphRAG, gpt4o & 48 \\
\bottomrule
\end{tabular}
}
% \caption{MINE information-retention scores (\S\ref{sec:eval-mine}). AutoOnto's hierarchical structure improves disambiguation and hence retrieval.}
\caption{MINE-1 information retention: the fraction of facts supported by context retrieved from the constructed KG. With GPT-4.1 as the construction backbone, AutoOntic achieves 85.87\% without a supplied ontology, exceeding KGGen and Wikontic. Initializing AutoOntic with Wikidata seed classes further increases retention to 88.00\%.
% $D$ denotes the maximum induced hierarchy depth.
}
\label{tab:mine}
\end{table}

\section{Results}
\label{sec:results}

\subsection{Text2KGBench}
\label{sec:results-text2kgbench}

Table~\ref{tab:text2kgbench-types} reports the domain/range
type-judgment distribution (\S\ref{sec:evaluation}) over matched pairs, separately for each ontology subset. It measures how well the induced ontology's classes align with the
reference one. On Wikidata-TekGen, the induced type is an exact match to the gold class
53.6--58.6\% of the time; where it differs, the remainder splits between a
broader (\textsc{More General}) and a narrower (\textsc{More Narrow})
induced type rather than an unrelated one (\textsc{Not Related} stays
under 10\% on the domain side and under 8\% on the range side for both backbones). These results provide evidence that most
disagreement with the reference ontology is a matter of granularity, not an unrelated branch of the induced hierarchy.

For DBpedia-WebNLG, exact match is lower (27.3--42.4\% across sides and backbones) and \textsc{Not Related} is
higher (11.3--12.2\%, versus under 10\% on Wikidata-TekGen). Even so, granularity mismatches (\textsc{More General}/\textsc{More Narrow})
still account for roughly 80\% of non-exact matches on both sides, so the
induced hierarchy is still placing DBpedia-WebNLG's types in
approximately the right region.

Constraint Plausibility depends on the hierarchy because AutoOntic derives each relation's domain and range constraints by generalizing observed argument types to their lowest common ancestors (LCAs). A single overly broad type can therefore make a relation's constraint less specific, even when most instances have informative types. This issue appears more pronounced on Wikidata-TekGen: the raw type vocabulary is substantially more fragmented relative to the gold classes than on DBpedia-WebNLG (~\ref{tab:text2kgbench-types}). Seeding the hierarchy with Wikidata classes can anchor its upper levels, but does not resolve fragmentation throughout the hierarchy. Limiting hierarchy depth has a larger effect in several conditions, often increasing the share of constraints classified as related to gold classes. However, this increase can reflect coarser constraints rather than better alignment: a broad ancestor may turn a \textsc{Not Related} judgment into \textsc{More General} without increasing exact matches. These results show that Constraint Plausibility is sensitive to hierarchy structure and should not be interpreted as a standalone measure of ontology quality.
% \subsection{DBpedia-WebNLG}
% \label{sec:results-dbpedia}

% Table~\ref{tab:dbpedia-webnlg-separate} reports recall and extra rate for
% both backbones on all 19 DBpedia-WebNLG domains. Unlike
% Table~\ref{tab:text2kgbench-separate}, these figures are not passed
% through a groundedness filter: DBpedia-WebNLG's sentences are generated
% directly from the gold triples (WebNLG's own construction method) rather
% than sampled from Wikipedia article bodies, so we did not perform a
% groundedness check for this subset. Averaged recall is 78.8\% for
% gpt-oss-120b and 76.2\% for Qwen3-Next-80B, with extra rates of 23.0\%
% and 25.1\% respectively -- the two backbones perform much more similarly
% here than on Wikidata-TekGen's domain/range type-judgment
% (Table~\ref{tab:text2kgbench-types}), where the gap between them was
% substantially wider.

% Table~\ref{tab:text2kgbench-types}'s DBpedia-WebNLG rows report the same
% type-judgment distribution for this subset. Exact match is lower here
% (27.2--41.0\% across sides and backbones) and \textsc{Not Related} is
% higher (10.9--12.9\%, versus under 9\% on Wikidata-TekGen). Even so, granularity mismatches (\textsc{More General}/\textsc{More Narrow})
% still account for roughly 80\% of non-exact matches on both sides, so the
% induced hierarchy is still placing DBpedia-WebNLG's types in
% approximately the right region.

\subsection{MINE}
\label{sec:results-mine}

Table~\ref{tab:mine} reports reports information retention on MINE-1. AutoOntic's KG built with gpt-4.1 reaches 85.87\% accuracy vs KGGen 82.37\% and Wikontic 82.93. Initializing the hierarchy with Wikidata seed classes further raises retention to 88.00. Unlike the Text2KGBench evaluation above, MINE does not score agreement with a reference schema directly. It queries the constructed KG through embedding-similarity retrieval and a fixed-depth graph walk over entities. So this result demonstrates that the typed, canonicalized KG built for a corpus with no supplied schema improves downstream retrieval.

\paragraph{Ontology Quality.} Taken together, Tables~\ref{tab:text2kgbench-types}--\ref{tab:mine} suggest AutoOntic's schema-free construction recovers a usable typed ontology. Domain/range disagreement with the reference ontology (Table~\ref{tab:text2kgbench-types}) is dominated by granularity mismatches (\textsc{More General}/\textsc{More Narrow}) on both Text2KGBench subsets. Unrelated types (\textsc{Not Related}) stay under 10\% on Wikidata-TekGen and under 13\% on DBpedia-WebNLG, on both the domain and range sides. Together with MINE's retrieval accuracy, this is evidence that the induced hierarchy places types in approximately the right region of the ontology, and that this structure is usable for downstream retrieval, even where it does not converge on the exact reference granularity.

\section{Conclusion}
We presented AutoOntic, a schema-free pipeline that extends KG extraction and canonicalization with a hierarchical type ontology, hierarchy-aware entity deduplication, and induced domain/range constraints. On Text2KGBench, over 90\% of induced types in  are semantically aligned with reference classes, with most mismatches reflecting differences in granularity rather than unrelated types. On MINE-1, AutoOntic achieves 85.87\% information-retention accuracy, increasing to 88.00\% with Wikidata-seeded hierarchy initialization, outperforming KGGen and Wikontic. These results demonstrate that schema-free ontology induction can recover useful semantic structure and support effective downstream retrieval.

% Each stage targets a specific failure mode of naively applying flat deduplication to typed structure: conflating synonymy with hyponymy, letting placement order corrupt the resulting hierarchy, and letting relation canonicalization silently merge a predicate with its own grammatical inverse.

\section*{Limitations}

Our evaluation measures agreement between the induced ontology and a benchmark's reference schema, and the usability of the resulting KG for downstream retrieval; it does not directly measure extraction precision, since a schema-free pipeline can correctly extract facts a narrow, curated reference ontology simply does not cover, and such extractions are excluded from recall by design rather than scored as errors. Finally, our evaluation covers English-language corpora and existing benchmark ontologies only; generalization to other languages remains untested.

% \section*{Acknowledgments}
% TODO

\bibliography{custom}

@article{lewis2020retrieval,
  title={Retrieval-augmented generation for knowledge-intensive nlp tasks},
  author={Lewis, Patrick and Perez, Ethan and Piktus, Aleksandra and Petroni, Fabio and Karpukhin, Vladimir and Goyal, Naman and K{\"u}ttler, Heinrich and Lewis, Mike and Yih, Wen-tau and Rockt{\"a}schel, Tim and others},
  journal={Advances in neural information processing systems},
  volume={33},
  pages={9459--9474},
  year={2020}
}

@article{edge2024graphrag,
  title={From local to global: A graph rag approach to query-focused summarization},
  author={Edge, Darren and Trinh, Ha and Cheng, Newman and Bradley, Joshua and Chao, Alex and Mody, Apurva and Truitt, Steven and Metropolitansky, Dasha and Ness, Robert Osazuwa and Larson, Jonathan},
  journal={arXiv preprint arXiv:2404.16130},
  year={2024}
}

@inproceedings{mo2025kggen,
 author = {Mo, Belinda and Yu, Kyssen and Kazdan, Joshua and Mpala, Proud and Yu, Lisa and Kanatsoulis, Charilaos and Koyejo, Sanmi},
 booktitle = {Advances in Neural Information Processing Systems},
 editor = {D. Belgrave and C. Zhang and H. Lin and R. Pascanu and P. Koniusz and M. Ghassemi and N. Chen},
 pages = {30092--30115},
 publisher = {Curran Associates, Inc.},
 title = {{KGGen}: Extracting Knowledge Graphs from Plain Text with Language Models},
 url = {https://proceedings.neurips.cc/paper_files/paper/2025/file/2b368455e832d2b1a60bcad8c4c6481f-Paper-Conference.pdf},
 volume = {38},
 year = {2025}
}

@inproceedings{campello2013hdbscan,
  title={Density-based clustering based on hierarchical density estimates},
  author={Campello, Ricardo JGB and Moulavi, Davoud and Sander, J{\"o}rg},
  booktitle={Pacific-Asia conference on knowledge discovery and data mining},
  pages={160--172},
  year={2013},
  organization={Springer}
}

@article{izacard2022contriever,
  title={Unsupervised dense information retrieval with contrastive learning},
  author={Izacard, Gautier and Caron, Mathilde and Hosseini, Lucas and Riedel, Sebastian and Bojanowski, Piotr and Joulin, Armand and Grave, Edouard},
  journal={arXiv preprint arXiv:2112.09118},
  year={2021}
}

@inproceedings{weeds2004characterising,
    title = "Characterising Measures of Lexical Distributional Similarity",
    author = "Weeds, Julie  and
      Weir, David  and
      McCarthy, Diana",
    booktitle = "{COLING} 2004: Proceedings of the 20th International Conference on Computational Linguistics",
    month = "aug 23–aug 27",
    year = "2004",
    address = "Geneva, Switzerland",
    publisher = "COLING",
    url = "https://aclanthology.org/C04-1146/",
    pages = "1015--1021"
}

@inproceedings{mihindukulasooriya2023text2kgbench,
  title={Text2kgbench: A benchmark for ontology-driven knowledge graph generation from text},
  author={Mihindukulasooriya, Nandana and Tiwari, Sanju and Enguix, Carlos F and Lata, Kusum},
  booktitle={International semantic web conference},
  pages={247--265},
  year={2023},
  organization={Springer}
}

@inproceedings{agarwal2021tekgen,
  title={Knowledge graph based synthetic corpus generation for knowledge-enhanced language model pre-training},
  author={Agarwal, Oshin and Ge, Heming and Shakeri, Siamak and Al-Rfou, Rami},
  booktitle={Proceedings of the 2021 Conference of the North American Chapter of the Association for Computational Linguistics: Human Language Technologies},
  pages={3554--3565},
  year={2021}
}

@article{qwen2025qwen3,
  title={Qwen3 technical report},
  author={Yang, An and Li, Anfeng and Yang, Baosong and Zhang, Beichen and Hui, Binyuan and Zheng, Bo and Yu, Bowen and Gao, Chang and Huang, Chengen and Lv, Chenxu and others},
  journal={arXiv preprint arXiv:2505.09388},
  year={2025}
}

@article{openai2025gptoss,
  title={gpt-oss-120b \& gpt-oss-20b model card},
  author={Agarwal, Sandhini and Ahmad, Lama and Ai, Jason and Altman, Sam and Applebaum, Andy and Arbus, Edwin and Arora, Rahul K and Bai, Yu and Baker, Bowen and Bao, Haiming and others},
  journal={arXiv preprint arXiv:2508.10925},
  year={2025}
}

@inproceedings{chepurova2026wikontic,
    title = "Wikontic: Constructing {W}ikidata-Aligned, Ontology-Aware Knowledge Graphs with Large Language Models",
    author = "Chepurova, Alla  and
      Bulatov, Aydar  and
      Burtsev, Mikhail  and
      Kuratov, Yuri",
    editor = "Demberg, Vera  and
      Inui, Kentaro  and
      Marquez, Llu{\'i}s",
    booktitle = "Proceedings of the 19th Conference of the {E}uropean Chapter of the {A}ssociation for {C}omputational {L}inguistics (Volume 1: Long Papers)",
    month = mar,
    year = "2026",
    address = "Rabat, Morocco",
    publisher = "Association for Computational Linguistics",
    url = "https://aclanthology.org/2026.eacl-long.388/",
    doi = "10.18653/v1/2026.eacl-long.388",
    pages = "8304--8319",
    ISBN = "979-8-89176-380-7"
}

\newpage
\appendix

\section{Hyperparameters}
\label{sec:appendix-hyperparams}

Table~\ref{tab:hyperparams} lists every hyperparameter referenced by symbol in \S\ref{sec:methods}, together with the value used for both reported backbones (Qwen3-Next-80B-A3B-Instruct and gpt-oss-120b, \S\ref{sec:evaluation}), which are run with an identical configuration apart from the LLM endpoint itself.

\begin{table*}[t]
\centering
\small
\begin{tabular}{llr}
\toprule
Symbol & Description & Value \\
\midrule
\multicolumn{3}{l}{\textit{Relation deduplication (\S\ref{sec:relation-dedup})}} \\
$\theta_r$ & HDBSCAN cosine-similarity threshold & 0.75 \\
$R_r$ & Merge-round cap & 5 \\
\addlinespace
\multicolumn{3}{l}{\textit{Entity-type canonicalization (\S\ref{sec:type-dedup})}} \\
$\theta_t$ & HDBSCAN cosine-similarity threshold & 0.75 \\
$R_t$ & Merge-round cap & 5 \\
\addlinespace
\multicolumn{3}{l}{\textit{Hierarchy induction (\S\ref{sec:hierarchy-induction})}} \\
$B$ & Candidate batch size & 30 \\
$D_{\max}$ & Maximum placement depth & 50 \\
-- & Relation-argument weighting scheme & PPMI \\
\addlinespace
\multicolumn{3}{l}{\textit{Entity deduplication (\S\ref{sec:entity-dedup})}} \\
$\theta_e$ & HDBSCAN cosine-similarity threshold (per partition) & 0.85 \\
$R_e$ & Merge-round cap (per partition) & 5 \\
$C_e$ & Cluster-size split/stitch cap & 40 \\
\addlinespace
\multicolumn{3}{l}{\textit{Constraint induction (\S\ref{sec:constraints})}} \\
$\tau_{\text{hard}}$ & HARD confidence threshold & 0.90 \\
$\tau_{\text{soft}}$ & SOFT confidence threshold & 0.50 \\
$\tau_{\text{hint}}$ & HINT confidence threshold (discard below) & 0.20 \\
\addlinespace
\multicolumn{3}{l}{\textit{Shared across all stages}} \\
-- & Sentence embedding model & Contriever \\
-- & Embedding batch size & 64 \\
-- & Parallel LLM-verification workers & 8 \\
\bottomrule
\end{tabular}
\caption{Pipeline hyperparameters. All HDBSCAN thresholds $\theta$ are cosine-similarity thresholds; HDBSCAN's own \texttt{cluster\_selection\_epsilon} is set to $1-\theta$. The parallel-worker count affects wall-clock time only -- cluster-verification order is reassembled deterministically regardless of completion order (\S\ref{sec:reoccurring-design}).}
\label{tab:hyperparams}
\end{table*}

\section{Hierarchy induction example}
\begin{figure*}[t]
\centering
\includegraphics[width=\linewidth]{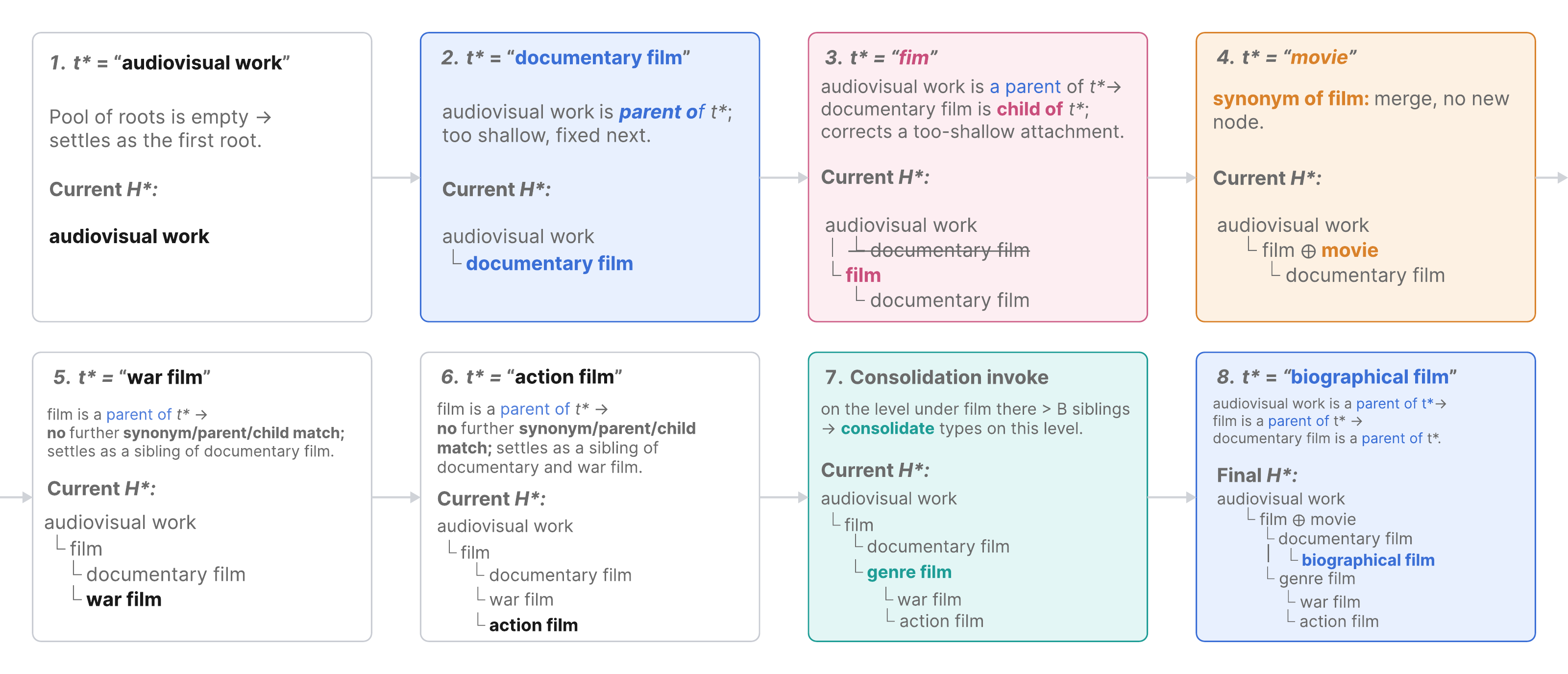}
\caption{Hierarchy induction on a film-domain vocabulary, one placement decision per step. (1) The initial $\mathcal{H}^*$ is empty, \emph{audiovisual work} settles as a root; (2) once \emph{documentary film} is placed, LLM judges \emph{audiovisual work} as a parent of \emph{documentary film}, which is too shallow; (3) LLM judges \emph{audiovisual work} as a parent of \emph{film} and \emph{documentary film} as a child of \emph{film}, reparenting too shallow is-a relation; (4) \emph{movie} is identified as a synonym of film; (5) \emph{war film} settles under \emph{audiovisual work} and then under \emph{film} as a child; (6) \emph{action film} is judged as a child of \emph{audiovisual work} and then as a child of \emph{film}; (7) there are too many siblings under \emph{film}: consolidating war and action film under a common parent, \emph{genre film}; (h) \emph{biographical film} settles under \emph{documentary film}, giving the final tree.}
\label{fig:hierarchy-steps-example}
\end{figure*}

Figure~\ref{fig:hierarchy-steps-example} demonstrates the process of building a hierarchy on a toy example from the movie domain step by step. 

\section{Token Usage}
\label{sec:appendix-tokens}

Table~\ref{tab:appendix-tokens-text2kg} reports total prompt+completion
tokens consumed by extraction (Step 1, \S\ref{sec:extraction}) and the
five subsequent pipeline stages (\S\ref{sec:methods}), summed across all
domains, for each Text2KGBench condition reported in
\S\ref{sec:results-text2kgbench}.
% Table~\ref{tab:appendix-tokens-mine} reports the same breakdown for the
% three MINE conditions in \S\ref{sec:results-mine}; the two gpt-oss-120b
% rows there reuse a single shared extraction pass, since they differ only
% in \texttt{max\_depth} over the same extracted corpus
% (\S\ref{sec:eval-mine}). Token counts are read directly from each run's
% own usage accounting (\S\ref{sec:reoccurring-design}); judge/evaluation
% token usage is not tracked by our evaluation scripts and is not included
% here.

\begin{table}[t]
\centering
\small
\begin{tabular}{lrrr}
\toprule
Condition & Extraction & Pipeline\\
\midrule
TekGen / Qwen3-Next-80B    & 4.71 & 3.84 \\
TekGen / gpt-oss-120b      & 7.02 & 5.63\\
WebNLG / Qwen3-Next-80B     & 2.20 & 1.09\\
WebNLG / gpt-oss-120b       & 2.99 & 1.71\\
\bottomrule
\end{tabular}
\caption{Total tokens (prompt+completion, Millions), summed across all
domains, per Text2KGBench condition.}
\label{tab:appendix-tokens-text2kg}
\end{table}

% \begin{table}[t]
% \centering
% \small
% \begin{tabular}{lrrr}
% \toprule
% Condition & Extraction (M) & Pipeline (M) & Total (M) \\
% \midrule
% gpt-oss-120b, \texttt{max\_depth=50} & 0.44 & 2.13 & 2.57 \\
% gpt-oss-120b, \texttt{max\_depth=10} & 0.44 & 2.01 & 2.45 \\
% \bottomrule
% \end{tabular}
% \caption{Total tokens (prompt+completion, millions) for the three MINE
% conditions in \S\ref{sec:results-mine}. The two gpt-oss-120b rows share
% the same 0.44M-token extraction pass rather than each paying for it
% independently.}
% \label{tab:appendix-tokens-mine}
% \end{table}

\section{Per-Domain Recall and Extra Rate}
\label{sec:appendix-per-domain}

Tables~\ref{tab:text2kgbench-separate}--\ref{tab:dbpedia-webnlg-separate} report recall and extra rate
(\S\ref{sec:evaluation}) per domain. For Wikidata-TekGen subset, we measure quality only on the dataset items
whose gold triple is actually supported by its own sentence text, as
judged by an independent GPT-5.1 groundedness check
(Appendix~\ref{sec:appendix-groundedness}): Text2KGBench's test sentences
are drawn from mid-article Wikipedia text without their source article's
topic, so a substantial fraction of gold subjects are established only
anaphorically, outside what the pipeline is actually shown. 

\begin{table*}[t]
\centering
\small
\begin{tabular}{lcccc}
\toprule
& \multicolumn{2}{c}{Recall (\%, grounded-only)} & \multicolumn{2}{c}{Extra rate (\%)} \\
\cmidrule(lr){2-3} \cmidrule(lr){4-5}
Domain & Qwen3-Next-80B & gpt-oss-120b & Qwen3-Next-80B & gpt-oss-120b \\
\midrule
ont\_1\_movie     & 77.4 & 73.7 & 74.5 & 78.9 \\
ont\_2\_music     & 72.6 & 74.2 & 74.1 & 77.6 \\
ont\_3\_sport     & 44.3 & 46.5 & 87.6 & 87.2 \\
ont\_4\_book      & 72.5 & 74.5 & 78.4 & 80.6 \\
ont\_5\_military  & 52.9 & 63.0 & 90.7 & 90.0 \\
ont\_6\_computer  & 79.5 & 81.3 & 73.2 & 73.9 \\
ont\_7\_space     & 75.5 & 73.4 & 70.0 & 73.7 \\
ont\_8\_politics  & 83.2 & 78.3 & 81.7 & 83.4 \\
ont\_9\_nature    & 59.1 & 66.3 & 88.8 & 88.6 \\
ont\_10\_culture  & 43.4 & 54.7 & 89.9 & 89.9 \\
\midrule
Averaged & 68.9 & 69.9 & 79.6 & 81.5 \\
\bottomrule
\end{tabular}
\caption{Recall (grounded-only, see \S\ref{sec:results-text2kgbench}) and extra rate (\S\ref{sec:evaluation}) per Wikidata-TekGen domain, judged by GPT-5.1.}
\label{tab:text2kgbench-separate}
\end{table*}

\begin{table*}[t]
\centering
\small
\begin{tabular}{lcccc}
\toprule
& \multicolumn{2}{c}{Recall (\%, unfiltered)} & \multicolumn{2}{c}{Extra rate (\%)} \\
\cmidrule(lr){2-3} \cmidrule(lr){4-5}
Domain & Qwen3-Next-80B & gpt-oss-120b & Qwen3-Next-80B & gpt-oss-120b \\
\midrule
ont\_1\_university            & 71.4 & 71.0 & 29.1 & 23.9 \\
ont\_2\_musicalwork           & 75.5 & 79.5 & 28.4 & 31.2 \\
ont\_3\_airport               & 79.3 & 78.5 & 26.3 & 25.3 \\
ont\_4\_building              & 67.3 & 71.5 & 34.0 & 22.7 \\
ont\_5\_athlete               & 86.3 & 92.3 & 12.5 & 7.7  \\
ont\_6\_politician            & 80.2 & 79.2 & 15.8 & 17.4 \\
ont\_7\_company               & 75.8 & 83.4 & 18.5 & 13.2 \\
ont\_8\_celestialbody         & 90.1 & 92.4 & 7.8  & 10.0 \\
ont\_9\_astronaut             & 64.9 & 73.5 & 31.7 & 25.7 \\
ont\_10\_comicscharacter      & 82.2 & 80.4 & 10.2 & 18.1 \\
ont\_11\_meanoftransportation & 77.5 & 77.5 & 16.7 & 18.3 \\
ont\_12\_monument             & 58.2 & 63.6 & 52.9 & 48.5 \\
ont\_13\_food                 & 71.4 & 67.3 & 34.7 & 42.4 \\
ont\_14\_writtenwork          & 81.9 & 79.8 & 16.4 & 17.2 \\
ont\_15\_sportsteam           & 74.8 & 79.6 & 17.8 & 13.3 \\
ont\_16\_city                 & 64.5 & 63.3 & 43.2 & 38.2 \\
ont\_17\_artist               & 71.8 & 75.8 & 27.9 & 23.3 \\
ont\_18\_scientist            & 86.1 & 87.8 & 12.2 & 12.6 \\
ont\_19\_film                 & 92.3 & 93.7 & 16.1 & 17.3 \\
\midrule
Pooled & 76.4 & 77.9 & 24.9 & 23.9 \\
\bottomrule
\end{tabular}
\caption{Recall and extra rate (\S\ref{sec:evaluation}) per DBpedia-WebNLG domain, judged by GPT-5.1.}
\label{tab:dbpedia-webnlg-separate}
\end{table*}

\section{Sentence Groundedness Analysis}
\label{sec:appendix-groundedness}

Text2KGBench's Wikidata-TekGen sentences are drawn from mid-article Wikipedia text without their source article's title or topic, while its gold triplets are annotated against the full article. The underlying corpus's own construction methodology \citep{agarwal2021tekgen} aligns a Wikidata triple to a sentence by requiring only the \emph{object} entity's alias to occur in it; when the \emph{subject} is absent, a single third-person pronoun is substituted with the subject's canonical name -- but this does not cover zero-anaphora (topic continuity with no pronoun to replace at all), which is the pattern we observe empirically in most cases. To our knowledge this has not been previously quantified specifically for Text2KGBench.
%This class of noise is well documented for distant supervision in relation extraction generally \citep{mintz2009distant, riedel2010modeling}, and DocRED \citep{yao2019docred} reports an analogous statistic built to motivate the opposite fix (document-level rather than sentence-level extraction): at least 40.7\% of its relational facts require reading multiple sentences. 
% though it is consistent with the known limitations of the distant-supervision alignment its underlying corpus uses.

We use an additional GPT-5.1 judge to classify, per gold triple, whether its sentence alone -- no outside knowledge, no article context -- actually supports it: the subject must be identifiable within the sentence (allowing name-form variation, e.g. ``Charles Spurgeon Johnson'' in the sentence vs.\ ``Charles S. Johnson'' in the reference still counts as grounded), and the object and relationship must actually be stated or implied. Recall in \S\ref{sec:results-text2kgbench} is reported with flagged (ungrounded) triplets excluded from the denominator, as a post-filter over the already-collected judge results requiring no re-extraction or re-matching.

Table~\ref{tab:text2kgbench-grounded} reports recall before and after this filter, per domain (Qwen3-Next-80B): Averaged across Wikidata-TekGen, 48.8\% of gold triplets are flagged ungrounded. Table~\ref{tab:text2kgbench-grounded-split} makes the comparison more direct -- recall computed \emph{separately} over the grounded and the ungrounded subsets, Averaged across all ten domains, for both backbones. Recall on grounded triplets is more than three times that on ungrounded ones for both models, evidence the classification tracks something real rather than adding noise on top of noise: a triple the judge calls grounded is recovered at roughly the rate the pipeline's own extraction and matching quality would predict, while an ungrounded one is only matched when the extractor happens to supply outside knowledge the sentence alone did not license.

% For reference,
% recall computed over the full, unfiltered set of gold triplets -- 45.7\%
% (Qwen3-Next-80B) and 46.1\% (gpt-oss-120b) -- is already comparable to, or
% roughly 0.1 points above, the best baseline (Vicuna-13B, recall 0.34)
% reported in the original Text2KGBench study on this same subset
% \citep{mihindukulasooriya2023text2kgbench}, though the two settings are
% not directly comparable: the original task supplies the gold ontology and
% asks for ontology-compliant extraction, while AutoOntic receives no
% ontology and induces its own schema from scratch.
\begin{table}[t]
\centering
\small
\setlength{\tabcolsep}{3pt}
\resizebox{0.8\columnwidth}{!}{%
\begin{tabular}{lccc}
\toprule
Domain & Raw & Filtered & Ungr. (\%) \\
\midrule
ont\_1\_movie     & 42.9 & 77.4 & 62.2 \\
ont\_2\_music     & 46.9 & 72.6 & 51.2 \\
ont\_3\_sport     & 32.0 & 44.3 & 32.1 \\
ont\_4\_book      & 48.6 & 72.5 & 45.8 \\
ont\_5\_military  & 39.1 & 52.9 & 38.7 \\
ont\_6\_computer  & 59.8 & 79.5 & 34.5 \\
ont\_7\_space     & 69.9 & 75.5 & 13.6 \\
ont\_8\_politics  & 67.8 & 83.2 & 29.2 \\
ont\_9\_nature    & 36.9 & 59.1 & 48.0 \\
ont\_10\_culture  & 34.1 & 43.4 & 69.4 \\
\midrule
Averaged & 45.4 & 68.9 & 48.8 \\
\bottomrule
\end{tabular}%
}
\caption{Recall (\%) before and after excluding gold triplets an independent GPT-5.1 judge flags as not grounded in their own sentence, and the Averaged fraction of gold triplets so flagged (Qwen3-Next-80B, Wikidata-TekGen). The Averaged ungrounded rate (48.8\%) differs from the per-domain average because it is computed over triplets, not domains.}
\label{tab:text2kgbench-grounded}
\end{table}

\begin{table}[t]
\centering
\small
\resizebox{\columnwidth}{!}{%
\begin{tabular}{lcc}
\toprule
Model & Recall, grounded & Recall, ungrounded \\
\midrule
Qwen3-Next-80B & 68.9 & 20.6 \\
gpt-oss-120b   & 69.9 & 19.7 \\
\bottomrule
\end{tabular}
}
\caption{Recall (\%) computed separately over the grounded and ungrounded gold-triple subsets, Averaged across all ten Wikidata-TekGen domains, \textsc{Separate} condition ($n=3{,}705$ grounded / $3{,}527$ ungrounded triplets, both models).}
\label{tab:text2kgbench-grounded-split}
\end{table}

\paragraph{Examples.} \emph{Ungrounded -- subject established anaphorically, not in the sentence:} the sentence \emph{``The series depicted the Great Migration of African Americans from the rural South to the urban North.''} is paired with the gold triple (\emph{Jacob Lawrence}, \emph{ethnic group}, \emph{African Americans}); \emph{Jacob Lawrence} is the subject of the source Wikipedia article but is never named in this particular sentence, so no extractor shown only this sentence could recover the fact.

\emph{Grounded -- name-form variation, not omission:} the sentence \emph{``Charles Spurgeon Johnson (July 24, 1893 -- October 27, 1956) was an American sociologist and college administrator, the first black president of historically black Fisk University, and a lifelong advocate for racial equality and the advancement of civil rights for African Americans and all ethnic minorities.''} is paired with the gold triple (\emph{Charles S. Johnson}, \emph{ethnic group}, \emph{African Americans}). The sentence names the subject in a different but recognizable form and states the object explicitly, so the judge correctly marks it grounded rather than flagging it for omission.

\section{Prompt Templates}
\label{sec:appendix-prompts}

This appendix lists, verbatim, the system prompt template behind each of
the eight LLM calls the orchestrated pipeline makes, without full example's texts for simplicity (full prompts are shipped in
\texttt{src/ontodisco/utils/prompts/}.

\subsection{Typed Triplet Extraction}
\label{sec:appendix-extraction}
Used by the schema-free extractor (\S\ref{sec:extraction}) to produce the raw triplet stream $\mathcal{G}$ that every downstream stage operates on.

\begin{lstlisting}[caption={System prompt for typed triplet extraction.}, label={lst:extraction}]
You are an algorithm designed to extract structured knowledge from texts to build a Wikidata-like knowledge graph. A knowledge graph consists of **triplets** in the format (subject, relation, object), where:

- **Subject**: A named entity or a concept that describes a group of people, events, or any abstract objects that serves as the source of the relation.
- **Relation**: A Wikidata-style predicate that connects the subject and object.
- **Object**: A named entity or a concept that describes a group of people, events, or any abstract objects that is related to the subject.

Additionally, some triplets may have **qualifiers** that provide more context (e.g., date, place, or other attributes). Qualifiers should have relations and object like triplets do, but instead of subject their relation connects an object and the triplet qualifier belongs to. **Qualifiers must always be attached to a triplet** and never exist as standalone triplets.

You will receive a text labeled **"Text:"**. Your task is to extract meaningful triplets that represent factual relationships.

### Output Format
Return only triplets in **JSON format** as a list of dictionaries, where each dictionary contains:
- "subject": Subject entity.
- "relation": Relation connecting subject and object.
- "object": Object entity.
-  "qualifiers": List of dictionaries, where each dictionary contains:
    - "relation": Relation connecting triplet and object,
    - "object": Object entity connected to the main triplet
- "subject_type": a class that describes the subject
- "object_type": a class that describes the object

NO additional text, NO "```" in json, return ONLY triplets in JSON format.


<example>
...
</example>
\end{lstlisting}

\subsection{Relation Cluster Verification}
\label{sec:appendix-cluster-relations}
Used during relation canonicalization (\S\ref{sec:relation-dedup}) to adjudicate a merge/split decision for one HDBSCAN candidate cluster of raw relation surface forms.

\begin{lstlisting}[caption={System prompt for relation cluster merge/split verification.}, label={lst:cluster-relations}]
You are an ontology engineer. Below is a group of relation labels that have been clustered together based on semantic similarity.

Your task:
1. Decide whether ALL labels refer to the SAME conceptual relation.
2. If yes: choose the BEST canonical label (most specific, most common in the field).
3. If no: split the group into sub-groups of labels that DO refer to the same relation,  and for each sub-group give a canonical label.

Return ONLY valid JSON (no explanation, no markdown), using this format:

If merging all into one relation:
{"decision": "merge", "groups": [{"canonical_label": "...", "members": ["...", ...]}]}

If splitting into multiple relations:
{"decision": "split", "groups": [{"canonical_label": "...", "members": ["...", ...]}, ...]}"""
\end{lstlisting}

\subsection{Entity-Type Cluster Verification}
\label{sec:appendix-cluster-entity-types}
Used during type canonicalization (\S\ref{sec:type-dedup}) to adjudicate a merge/split decision for one candidate cluster of raw entity-type labels, including the hyponymy (``kind of'') test discussed in \S\ref{sec:type-dedup}.

\begin{lstlisting}[caption={System prompt for entity-type cluster merge/split verification, including the hyponymy guard.}, label={lst:cluster-entity-types}]
You are an ontology engineer. Below is a group of entity type labels that have been clustered together based on semantic similarity. Some candidates may be annotated with "(context: ...)", describing which relations they tend to appear as the subject/object of in the corpus.

Your task:
1. Decide whether ALL labels refer to the SAME conceptual type.
2. If yes: choose the BEST canonical label (most specific, most common in the field).
3. If no: split the group into sub-groups of labels that DO refer to the same type,  and for each sub-group give a canonical label.

CRITICAL TEST for step 1 -- two labels are the SAME type only if they are simply
two names for the identical concept, interchangeable in any sentence. Before
merging, apply this test: can you naturally say "an X is a kind of Y" (or "a Y,
specifically an X")? If so, they are NOT the same type -- X is a more specific
type and Y is a more general one, and they MUST be split into separate groups,
even though they are obviously related and even though they may share almost
all the same context relations. For example: "documentary film" is a kind of
"film" -> split, never merge, regardless of how similar their contexts look.
"movie" and "film" are just two names for the same thing, with no "kind of"
relationship between them -> merge. Heavy context/relation overlap is expected
and NORMAL for a specific type and its general type (a documentary film
participates in almost every relation "film" does) -- it is NOT by itself
evidence for merging; the "kind of" test always takes priority over context
similarity. When genuinely unsure whether two labels are identical or in a
"kind of" relationship, prefer splitting: a missed merge can still be corrected
later, but a wrong merge silently and permanently destroys the distinction.

Contexts remain useful evidence for the OPPOSITE case: if two labels look
similar but their contexts share no relations in common at all, prefer
splitting them into different (likely unrelated or homonymous) types.

Return ONLY valid JSON (no explanation, no markdown), using this format:

If merging all into one type:
{"decision": "merge", "groups": [{"canonical_label": "...", "members": ["...", ...]}]}

If splitting into multiple types:
{"decision": "split", "groups": [{"canonical_label": "...", "members": ["...", ...]}, ...]}"""
\end{lstlisting}

\subsection{Hierarchy Placement (Parent/Child Checks and Sibling Regrouping)}
\label{sec:appendix-hierarchy-pairwise}
Used by hierarchy induction (\S\ref{sec:hierarchy-induction}) to resolve how one focal type relates to the candidates at its current level. Rather than one combined prompt offering a parent/child/synonym/sibling/none choice, the decision is decomposed into two single-purpose calls -- a parent check, and (only if that finds nothing) a child check -- plus a separate sibling-regrouping call issued only when a level exceeds the candidate batch size $B$ (\S\ref{sec:hierarchy-induction}).

\begin{lstlisting}[caption={System prompt for the parent check: does the focal type belong under one of the candidates?}, label={lst:hierarchy-parent-check}]
You are an ontology engineer. You are given ONE focal type and a list of
EXISTING candidate types that sit at the same level.

Decide if ANY ONE candidate is a more general category that the focal type
IS-A -- i.e. "every <focal> is a kind of <candidate>" is literally true. This
is the ONLY question. Do not consider whether focal is a parent of any
candidate -- that is asked separately, in a different call.

These are NOT "is-a", no matter how strongly associated the two are:
- focal happens DURING / emerged in / belongs to the era of the candidate
- focal is PART OF, made of, or located in the candidate
- focal is CAUSED BY, produced by, used by, or studied by the candidate
- focal is merely a related topic, or co-occurs with the candidate

If a candidate is not just a broader category but the SAME concept as focal
(a true synonym, e.g. "vehicle" / "means of transport"), say so explicitly --
that is a merge, not a parent relationship.

Some types carry "(context: ...)" -- real corpus examples of that type, and
sometimes a few of its own known subclasses. Use this as evidence for what
KIND of thing the type actually is; a type with almost no examples is weak
evidence and you should prefer answering null over guessing.

Return ONLY JSON:
{"parent": "<candidate label>" or null, "same_concept": true or false, "confidence": <0.0-1.0>}

"same_concept" is only ever true when "parent" is non-null and they are the
identical concept, not merely related by generality.
\end{lstlisting}

\begin{lstlisting}[caption={System prompt for the child check: does the focal type subsume one or more candidates? Only asked if the parent check found nothing.}, label={lst:hierarchy-child-check}]
You are an ontology engineer. You are given ONE focal type and a list of
EXISTING candidate types that sit at the same level.

Decide which candidates, if any, the focal type is a more general category
OVER -- i.e. "every <candidate> is a kind of <focal>" is literally true for
that candidate. Apply this test to EACH candidate SEPARATELY.

These are NOT "is-a", no matter how strongly associated the two are:
- the candidate happens DURING / emerged in / belongs to the era of focal
- the candidate is PART OF, made of, or located in focal
- the candidate is CAUSED BY, produced by, used by, or studied by focal
- the candidate is merely a related topic, or co-occurs with focal

Genuine multi-candidate cases are real but usually small -- two or three
closely related candidates ("war film" and "action film" under "film
genre"). If you find yourself naming most of the list, you are answering a
looser question ("which of these relate to focal?") instead of this one --
stop, and re-apply the test to each candidate on its own.

Some types carry "(context: ...)" -- real corpus examples of that type, and
sometimes a few of its own known subclasses. Use this as evidence for what
KIND of thing the type actually is; a type with almost no examples is weak
evidence, and claiming it as a child on that basis alone is not warranted.

Return ONLY JSON:
{"children": ["<label>", ...], "confidence": <0.0-1.0>}

Empty list if none qualify. confidence applies to the set as a whole.
\end{lstlisting}

\begin{lstlisting}[caption={System prompt for sibling regrouping, issued only when a level's candidate count exceeds $B$.}, label={lst:hierarchy-regroup}]
You are an ontology engineer. You are given a list of types that all sit at
the SAME level -- none is a parent of another -- but there are too many to
compare one at a time, so you must group them first.

For each group of 2 or more types that plausibly share a common, currently
NONEXISTENT parent category (true siblings -- e.g. "war film" and "action
film" both being kinds of film genre), invent ONE short canonical label and a
one-sentence definition for that shared parent.

Rules:
- Only group true SIBLINGS -- same level of abstraction, same kind of thing.
  Do not group merely-related types, and do not group a general type with a
  specific one (that is a parent/child relationship, not a sibling group --
  leave both out of any group).
- A type may appear in AT MOST ONE group.
- Any type that does not clearly belong in a group: leave it out entirely.
  Precision matters more than coverage -- when in doubt, leave it out.
- Before inventing a label, check whether an existing, more general word
  already used elsewhere would describe the group just as well. If so,
  prefer reusing a short, ordinary word over a novel compound.

Return ONLY JSON:
{"groups": [{"label": "...", "definition": "...", "members": ["<exact input label>", ...]}, ...]}

Empty list if nothing groups. Every member list must have length >= 2.
\end{lstlisting}

\subsection{Entity-Name Cluster Verification}
\label{sec:appendix-cluster-entity-names}
Used during entity deduplication (\S\ref{sec:entity-dedup}) to adjudicate a merge/split decision for one candidate cluster of compound \texttt{name~[type]} entity mentions, within one parent-gated partition.

\begin{lstlisting}[caption={System prompt for entity-name cluster merge/split verification.}, label={lst:cluster-entity-names}]
You are a knowledge graph engineer. Below is a group of entity mentions, each in the format "name [type]". These were clustered together by embedding similarity ONLY -- being in this list is NOT evidence that they are the same entity. Most clusters contain several DIFFERENT entities that merely sound alike or share a type.

Your task:
1. Decide which mentions are SYNONYMOUS: different ways of writing the SAME single real-world entity.
2. Group synonymous entities together. A mention with no synonym forms its own group of one.
3. For each group, choose the BEST canonical name -- the most complete, unambiguous form of that entity's name, taken from the mentions themselves. The canonical name must be JUST the entity name, WITHOUT the [type] bracket.
4. Each group's "members" array must contain the EXACT compound strings as provided (including the [type] part).

### The test for a duplicate

Two mentions are SYNONYMOUS ONLY if they are denoting the same person, organization, place, work, event, or concept -- such that **every statement true of one is true of the other**, and merging them would LOSE NO INFORMATION.

Apply it by substitution: if you swapped one mention for the other in a sentence, would the meaning stay identical?
- "Chris Nolan" / "Christopher Nolan" -- identical meaning. SYNONYMOUS.
- "Nobel Prize" / "Nobel Prize in Physics" -- one is the award family, the other a specific award. Merging discards "in Physics". NOT SYNONYMOUS.

### Type rules

- SIMILAR types allow merging only if the entity name, combined with the type, refers to the same real-world entity ("Christopher Nolan [film director]" / "Chris Nolan [filmmaker]" / "Nolan [human]").
- Abbreviations, nicknames, spelling and case variants, and partial names merge with their full form when the type context supports it.

### Do not invent a name

The canonical name must be one of the given mentions (or its fuller spelling). Never invent a broader umbrella term to cover a group -- if you find yourself reaching for a general word that covers several mentions, that is proof they are NOT duplicates and must stay separate.

When in doubt, keep entities SEPARATE. Precision matters far more than recall here: a wrong merge destroys two entities permanently, while a missed merge leaves both intact.

Return ONLY valid JSON (no explanation, no markdown). Every input mention must appear in exactly one group:
{"groups": [{"canonical_label": "<just the name>", "members": ["<name [type]>", ...]}, ...]}
\end{lstlisting}

\subsection{Relation-Direction Classification}
\label{sec:appendix-relation-direction}
Used by constraint induction's Stage~1 (\S\ref{sec:constraints}) to classify each raw surface form merged into a canonical relation as \emph{forward} or \emph{inverted} relative to that relation's own canonical direction, ahead of domain/range constraint mining.

\begin{lstlisting}[caption={System prompt for relation-direction (forward/inverted) classification.}, label={lst:relation-direction}]
You are an ontology engineer resolving relation directionality.

Below is a canonical relation (its chosen representative label) and the full list of raw
predicate surface forms that were merged into it during relation canonicalization. Surface forms
are merged when they denote the same underlying relationship between two entities, INCLUDING
cases where one form is the grammatical passive/inverse of another (e.g. "directed" and "was
directed by" describe the same relationship, just told with subject and object swapped).

Your task: classify EACH surface form relative to the canonical label's own direction, using
exactly one of:

1. "forward" -- this surface form keeps the same subject/object roles as the canonical label.
   Example: canonical "directed", surface form "helmed" -> "forward" (same roles: person directs
   film).

2. "inverted" -- this surface form swaps subject and object relative to the canonical label: what
   is the SUBJECT under the canonical label becomes the OBJECT under this surface form, and vice
   versa. Example: canonical "directed" (subject=director, object=film), surface form "was
   directed by" -> "inverted" (under this form, the film is the grammatical subject and the
   director the object).

The canonical label itself is always "forward" by definition -- it is the reference direction
everything else is judged against.

Base this purely on grammatical/semantic role structure, not on which surface form is more
frequent or "sounds more natural" -- frequency has nothing to do with direction.

Return ONLY valid JSON (no explanation, no markdown), using this format:
{
  "labels": {
    "directed": "forward",
    "was directed by": "inverted",
    "helmed": "forward"
  }
}

Every surface form given to you must appear as a key in "labels", with exactly one of "forward"
or "inverted" as its value. Do not invent surface forms that were not given to you.
\end{lstlisting}

\end{document}